\documentclass[sigconf]{acmart}
\AtBeginDocument{%
  \providecommand\BibTeX{{%
    \normalfont B\kern-0.5em{\scshape i\kern-0.25em b}\kern-0.8em\TeX}}}

\usepackage{mathtools}
\usepackage{todonotes}
\usepackage{siunitx}
\usepackage{cancel}
\usepackage{wrapfig}
\usepackage{balance}

\copyrightyear{2022} 
\acmYear{2022} 
\setcopyright{none}
\acmConference[NICE 2022]{Neuro-Inspired Computational Elements Conference}{March 28-April 1, 2022}{Virtual Event, USA}
\acmBooktitle{Neuro-Inspired Computational Elements Conference (NICE 2022), March 28-April 1, 2022, Virtual Event, USA}
\acmPrice{15.00}
\acmDOI{10.1145/3517343.3517368}
\acmISBN{978-1-4503-9559-5/22/03}
\begin{document}

\title[Integer Factorization with Distributed Representations]{Integer Factorization with Compositional Distributed Representations}

\author{Denis Kleyko}
\affiliation{%
  %\institution{University of California at Berkeley}
  \institution{UC Berkeley}
  \city{Berkeley}
  \state{CA}
  \country{USA}
  %\postcode{94720}
  \and
  \institution{Research Institutes of Sweden}
  \city{Kista}
  \country{Sweden}
  %\postcode{16440}  
}

\author{Connor Bybee}
\author{Christopher Kymn}
\author{Bruno Olshausen}
\affiliation{%
  %\institution{University of California at Berkeley}
  \institution{UC Berkeley}
  \city{Berkeley}
  \state{CA}
  \country{USA}
  %\postcode{94720}
}

\author{Amir Khosrowshahi}
\affiliation{%
  \institution{Technology Development, Intel}
  \city{Santa Clara}
  \state{CA}
  \country{USA}
  %\postcode{94720}
}

\author{Dmitri E. Nikonov}
\affiliation{%
  \institution{Components Research, Intel}
  \city{Hillsboro}
  \state{OR}
  \country{USA}
  %\postcode{94720}
}

\author{Friedrich T. Sommer}
\author{E. Paxon Frady}
\affiliation{%
  \institution{Intel Labs}
  \city{Santa Clara}
  \state{CA}
  \country{USA}
  %\postcode{95054}
}

\renewcommand{\shortauthors}{Kleyko, et al.}

\begin{abstract}
In this paper, we present an approach to integer factorization using distributed representations formed with Vector Symbolic Architectures.  
The approach formulates integer factorization in a manner such that it can be solved using neural networks and potentially implemented on parallel neuromorphic hardware. 
We introduce a method for encoding numbers in distributed vector spaces and explain how the resonator network can solve the integer factorization problem.
%The proposed approach is a combination of the log transformation and two primitives of Vector Symbolic Architectures: a fractional power encoding and a resonator network.
%
We evaluate the approach on factorization of semiprimes by measuring the factorization accuracy versus the scale of the problem.
% Want to put in  a clean result, I think something like this but double check if this makes sense
%We show that the size of the factorization problem that can be solved scales quadratically with network size.
We also demonstrate how the proposed approach generalizes beyond the factorization of semiprimes; in principle, it can be used for factorization of any composite number. 
%
%We hope that the proposed approach demonstrates how a well-known problem can be formulated and solved with compositional distributed representations and will inspire similar approaches to other difficult combinatorial search problems. 
This work demonstrates how a well-known combinatorial search problem may be formulated and solved within the framework of Vector Symbolic Architectures, and it opens the door to solving similarly difficult problems in other domains.
\end{abstract}



\maketitle

\section{Introduction}

Distributed information processing and distributed representations were proposed in the 1980s for solving optimization and factorization problems. For example, associative networks~\cite{hopfield1982neural} can solve optimization problems like the traveling salesman problem~\cite{hopfield1985neural}.
%There are several motivations for addressing optimization and factorization with distributed representations. 
This research direction continues to draw interest for many reasons. 
First, it can provide insights into how biological neural circuits solve optimization problems occurring in cognitive tasks. 
Second, distributed representations are well-suited for implementation in unconventional parallel hardware~\cite{JaegerComputing2020, MoughanHDParallel2021}, such as neuromorphic hardware~\cite{TrueNorth14, Loihi18, NNkNN20}, potentially providing scalable and low-energy solutions to challenging computing problems~\cite{DaviesAdvancingLoihi2021}.
%Other hardware types include quantum computers and coupled oscillators
%Third, there is recent interest in solving hard optimization problems by mapping the problem to a distributed representation, which can be solved efficiently on various unconventional hardware~
%\cite{jiang2018quantum, wang2019oim, borders2019integer, vadlamani2020physics}.
%Another favorable property of distributed representations is that they can be easily made executable within a parallel computing paradigm~\cite{}.

In this paper, we use a framework for forming structure-sensitive distributed representations that can flexibly encode compositional data structures~\cite{RachkovskijStructures2001} known as Vector Symbolic Architectures (VSA) a.k.a. Hyperdimensional Computing~\cite{KanervaHyperdimensional2009,GaylerJackendoff2003, FradySDR2020}.
In VSAs, the essential operation for forming distributed representations of compositional data structures is the binding operation~\cite{KleykoComputingParadigm2021}.
However, many VSA data structures require parsing, which amounts to a challenging combinatorial factorization problem that must be solved. 
%The simplest example of a compositional data structure formed by a single binding operation is a key-value pair.
%The use of the binding operation, however, creates a problem of recovering the individual representations used to obtain the result of the binding. 
%The binding operation also has an (approximate) inverse. 
%Given a key-value pair and the key, an unbinding with the key can be used to retrieve the value. However, if only the composite representation is given (without the key), it cannot be parsed in traditional VSA models.
Recently, resonator networks~\cite{FradyResonator2020} were proposed that can efficiently factor compositional representations into their constituents \cite{KentResonatorNetworks2020}. 
%The parsing is essentially a factorization of the compositional representation
%and results in~\cite{KentResonatorNetworks2020} suggest that the resonator network is very efficient in solving factorization problems. 
Here, we demonstrate how the VSA technique \emph{fractional power encoding} (FPE) \cite{PlateNested1994, FradyFunctions2021, FradyFunctionsNICE2022} can be used to represent integers as vectors and we show how the problem of factorizing integers can be expressed as the problem of factorizing vectors. We then explain how resonator networks can be extended to solve prime factorization of integers, and we measure the performance and scaling of our method.
%However, the problem formulation in~\cite{KentResonatorNetworks2020} is abstract, and it is not connected to any well-known factorization problem, so it is not obvious how to apply resonator networks for such problems. 
%Therefore, the goal of this paper is to provide a demonstration of how structure-sensitive distributed representations can be applied toward known factorization problems.
%To this end, the integer factorization is featured in this paper.
%\\

%The paper is structured as follows. 
%The main concepts are presented in Section~\ref{sect:concepts}. 
%Section~\ref{fact:setup} introduces the setup for integer factorization. 
%The empirical evaluation of the proposed approach is reported in Section~\ref{sect:empirical}. 
%The findings and their connection to related work are discussed in Section~\ref{sect:discussion}.
%Section~\ref{sect:conclusions} presents the concluding remarks.

\section{Methods}
\label{sect:concepts}

%There are three important ingredients used in the proposed distributed approach to integer factorization. The first is how to use VSAs to form distributed representations of data and data structures. The second is how fractional power encodings are used to represent numbers as vectors. And finally is the use of the resonator network to factorize vector representations.

% Hmm this is confusing that it doesnt match the subsections.
% \begin{itemize}

%     \item Fractional power encoding (FPE) in VSAs \cite{PlateNested1994,KomerContinuous2019, FradyFunctions2021}, which is a similarity-preserving encoding method that maps scalars $x$ to distributed representations $\mathbf{z}(x)$. Binding (denoted as $\odot$) of the FPE representations of two scalars produces the representation  
%     %combined with the proper binding operation , such that the result of binding of representations of two scalars corresponds to the representation 
%     of their sum: $\mathbf{z}(x+y)=\mathbf{z}(x) \odot \mathbf{z}(y)$;

%     \item The log transformation of scalars, which turns multiplication into addition, i.e., $\log(xy)=\log(x)+\log(y)$;

%     \item The resonator network~\cite{KentResonatorNetworks2020, FradyResonator2020}, which allows factorization of a compositional distributed representation of the form $\mathbf{a} \odot \mathbf{b} \odot \mathbf{c}$ into its constituents, $\mathbf{a}$, $\mathbf{b}$, and $\mathbf{c}$.
    
% \end{itemize}

%\section{Concepts}

\subsection{Vector Symbolic Architectures and Fourier Holographic Reduced Representations}
\label{sect:supp:vsa}

First in this section, we provide a brief overview of the required components from VSAs~\cite{KleykoComputingParadigm2021}. Please consult~\cite{KleykoSurveyVSA2021Part1,KleykoSurveyVSA2021Part2} for a comprehensive survey.
%using the Fourier Holographic Reduced Representations (FHRR) model to showcase a particular VSA realization. 
%It is important to keep in mind that VSA can be formulated with different types of vectors, namely those containing real, complex, or binary entries, as well as with the multivectors of geometric algebra.
%These VSA models come under many different names: Holographic Reduced Representation  (HRR)~\cite{PlateNested1994, PlateHolographic1995}, 
%Multiply-Add-Permute~\cite{GaylerMAP1998}, Binary Spatter Codes~\cite{KanervaFully1997}, Sparse Binary Distributed Representations~\cite{RachkovskijStructures2001,KleykoSDR2016}, Sparse Block-Codes~\cite{LaihoSparse2015, FradySDR2020}, Matrix Binding of Additive Terms~\cite{GallantRepresenting2013}, Geometric Analogue of Holographic Reduced Representation~\cite{AertsGeometric2009}, etc. 
%All of these different models have similar computational properties -- see~\cite{FradyCapacity2018} and~\cite{SchlegelVSAComparison2020}. 
%For a more in-depth introduction, we highly recommend consulting~\cite{KanervaHyperdimensional2009} and~\cite{NeubertRobotics2019}.
The key components of any VSA model are: a high-dimensional vector space where random vectors are pseudo-orthogonal ($n$ denotes the dimensionality); symbol representations with randomized atomic vectors (a.k.a. hypervectors; bold lowercase letters, e.g., $\mathbf{a}$); item memory for storing atomic hypervectors and performing auto-associative search (matrices indicated by bold uppercase letters, e.g., $\mathbf{A}$). 

% \noindent
% \begin{itemize}
%     \item High-dimensional vector space ($n$ denotes the dimensionality);
%     \item Atomic representations with randomized vectors (a.k.a. hypervectors);
%     \item Pseudo-orthogonality between random vectors;
%     \item Similarity measure (inner product or cosine similarity);
%     \item Item memory for storing atomic hypervectors and performing auto-associative search;
%     \item Operations on hypervectors.
% \end{itemize}
% \noindent

Here, we are utilizing a version of VSA known as Fourier Holographic Reduced Representations (FHRR)~\cite{PlateNested1994}.
In FHRR~\cite{PlateNested1994, PlateHolographic1995},
%was introduced by Plate as a model inspired by HRR~\cite{PlateAlgebra1991}. 
the atomic hypervectors are complex-valued random vectors, where each vector component can be considered as an angle (phasor) randomly and independently selected from the uniform distribution over $(0, 2\pi]$ and with magnitude of one.
The similarity measure is 
%the mean of cosines of angle differences of individual components of two hypervectors~\cite{PlateNested1994}. 
%Note that this measure can also be 
expressed by the normalized inner product between two phasor hypervectors ($\mathbf{a}$ and $\mathbf{b}$) as $
\frac{1}{n} \Re (\mathbf{a}^\dagger \mathbf{b})$, 
where $\mathbf{a}^{\dagger}$ is the complex conjugate transpose, and $\Re$ denotes the real part of the inner product.
Each VSA model defines key operations used to manipulate atomic hypervectors. In FHRR, these are: 
\emph{binding} (denoted as $\odot$), which is implemented as component-wise multiplication (Hadamard product); \emph{inverse and unbinding}, which in FHRR corresponds to taking the complex conjugate of the vector to unbind (inverse $\overline{\mathbf{a}}$/$\mathbf{a}^{-1}$) and applying the Hadamard product ($\mathbf{b} \odot \overline{\mathbf{a}}$); \emph{superposition}, (a.k.a. bundling, denoted as $+$) which is implemented as component-wise complex addition, possibly followed by some normalization function; and \emph{permutation}, which can be implemented through a convolution operation or permutation (denoted as $\rho$ but we do not use it here).
% \noindent
% \begin{itemize}
%     \item Binding (denoted as $\odot$; implemented as component-wise multiplication a.k.a. Hadamard product). The inverse of binding is called unbinding and implemented via binding with the hypervector conjugate;
%     \item Superposition a.k.a bundling (denoted as $+$; implemented as component-wise complex addition possibly followed by some normalization function);
%     \item Permutation (denoted as $\rho$; a cyclic-shift of components).
% \end{itemize}
% \noindent

% These operations are used in VSA to form compositional representations from atomic hypervectors. 
% Random i.i.d. vectors can serve as atomic hypervectors to represent ``symbols'' in VSA (i.e., categorical objects), since such vectors are pseudo-orthogonal to each other (due to the concentration of measure phenomenon) and, thus, are treated as dissimilar. 
% However, when input data are of a subsymbolic nature (e.g., numerical data), it is important to form their atomic hypervectors such that the similarity between hypervectors reflects the similarity between the original input data. 
% This can be achieved by generating atomic hypervectors in VSA using similarity-preserving encoding~\cite{palmetal94} (see, e.g., \cite{KussulRandom1999, RachkovskijVectors2005, RachkovskijScalars2005,RachkovskijSimilarityRP2015,RachkovskijFastDistance2017}). 
% Below, we introduce one particular type of similarity-preserving encoding known as FPE that is used in this paper. 

\subsection{Fractional Power Encoding}
\label{sec:fpe}

In standard VSA methods, randomized atomic vectors act as symbols and can be manipulated like traditional symbolic representations. 
However, when sub-symbolic data or continuous values need to be represented (e.g., to solve machine learning problems~\cite{RahimiBiosignal2019, GeClassificationReview2020,KleykoDensityEncoding2020}), it is important to form a similarity-preserving encoding. 
Fractional power encoding (FPE) is a method for such a similarity-preserving encoding, originally proposed in~\cite{PlateNested1994} (see Section 5.6) as a generalization of the fractional power vector~\cite{PlateRecurrent1992}.
This approach has recently received renewed interest for representing continuous manifolds, such as location in an environment \cite{KomerNavigation2020}, and has been connected to kernel methods for describing continuous functions \cite{FradyFunctions2021, FradyFunctionsNICE2022}.

The idea behind the fractional power encoding starts with a single atomic hypervector $\mathbf{z}$.
This vector can then be used to represent different integers through self-binding, where each self-binding step creates a new
hypervector that is dissimilar to all the others. For instance, the value of 2 is represented by $\mathbf{z} \odot \mathbf{z}$; 3 is represented by  $\mathbf{z} \odot \mathbf{z} \odot \mathbf{z}$, and so on.
This can be expressed as exponentiating the vector (component-wise) with the integer value, i.e. 
%in the HRR model the binding operation can be applied multiple times to the same  each time creating   
%For example, the fractional power vector can be used to represent integers, where the encoded integer, $i$, defines the number of times the base vector is bound to itself.
%In FHRR, where the binding operation is component-wise multiplication, 
the hypervector representation of integer $i$ is formed as: $\mathbf{z}(i) = \mathbf{z}^i$.

% \noindent
% \begin{equation}
%     \mathbf{z}(i) = \mathbf{z}^{i},
%     \label{eq:int_exp_binding}
% \end{equation}
% \noindent
It was recognized \cite{PlateNested1994} that this exponentiation process can be defined continuously when using complex-valued FHRR vectors. Thus, the same scheme can be easily generalized to encoding any scalar $x$ as: $\mathbf{z}(x) = \mathbf{z}^{\beta x}$, 
%i.e., self-binding reduces to the component-wise exponentiation using $i$ as the exponent of the phasors constituting the base vector $\mathbf{z}$. 
%Because the value of the exponent $i$ does not have to be an integer, the same scheme can be easily generalized to encoding any scalar $x$ as:
% \noindent
% \begin{equation}
%     \mathbf{z}(x) = \mathbf{z}^{\beta x},
%     \label{eq:fpe}
% \end{equation}
% \noindent
where we introduced a parameter $\beta$ that controls the similarity-preserving properties of the resulting encoding. 
%where $\beta$ denotes a bandwidth parameter, which controls width of the similarity kernel for the fractional power encoding. 
Hypervectors obtained with FPE preserve similarity between nearby values of $x$, while values further away have reduced similarity. The exact shape of this similarity metric defines the similarity kernel, and $\beta$ regulates the width of this similarity kernel.
%Fig.~\ref{fig:bandwidth} presents examples of similarity kernels obtained when the phasors of $\mathbf{z}$ are drawn uniformly from $[-\pi,\pi)$, and demonstrates that the choice of $\beta$ determines the width of the kernel.

\begin{comment}
\begin{figure}[tb]%[!ht]%[t!]
\centering
\includegraphics[width=1.0\columnwidth]{img/Example_bandwidth}
\caption{
Similarity kernels for FPE with base vectors sampled from uniform phase distributions. 
Panels illustrate the effect of the bandwidth parameter $\beta$ on the shape of the kernel. 
Gray and blue curves show the similarity kernels at two different locations $x=6.0$ and $x=10.0$, respectively to demonstrate the fact that the resulting similarity kernels are translation-invariant. 
The resulting similarity kernels approximate the sinc kernel.
The reported values are averages computed from $30$ simulation runs; $n=512$. 
The bars denote the standard deviations.
}
\label{fig:bandwidth}
\end{figure}
\end{comment}

%\subsubsection{Definition}
%\label{sec:fpe:def}

Another important property of FPE that we leverage is that it defines a systematic relationship between the binding operation and FPEs of scalars.
In particular, binding the hypervectors representing two scalars $x$ and $y$ results in a hypervector representing $x+y$:
\noindent
\begin{equation}
    \mathbf{z}(x) \odot \mathbf{z}(y) = \mathbf{z}^{\beta x} \odot \mathbf{z}^{\beta y} = \mathbf{z}^{\beta (x+y)} = \mathbf{z}(x+y).  
    \label{eq:fpe:sum}
\end{equation}

%Note that the resulting similarity kernels in Fig.~\ref{fig:bandwidth} highly resemble the sinc kernel.
%In fact, it can be proven that such an FPE converges to the perfect sinc kernel when $n \rightarrow \infty$~\cite{FradyFunctions2021}.
%It is worth noting that the shape of the similarity kernel can be chosen by selecting the probability density function used for sampling the phases for the base vector, which tightly connects the FPE method to random Fourier features~\cite{rahimi2007random}.
There is much more to be said about uses of FPE, including the representation of functions, the shape of the similarity kernel, and representations of multi-dimensional numerical data. We kindly refer interested readers to a recent thorough treatment of FPE in~\cite{FradyFunctions2021, FradyFunctionsNICE2022}. 

%\newpage

\subsection{Resonator Networks}
\label{sect:fac:rn}

In VSA, the representation of a conjunction of two or more hypervectors (e.g., $\mathbf{a}$, and $\mathbf{b}$) is achieved by binding: $\mathbf{s} = \mathbf{a}  \odot \mathbf{b}$.
The resulting hypervector $\mathbf{s}$ is pseudo-orthogonal to the argument hypervectors (factors), and every combination of arguments results in a unique $\mathbf{s}$.
The binding operation is invertible; when 
given all but one factor ($\mathbf{b}$), one can simply compute the unknown factor ($\mathbf{a}$) from the bound representation by unbinding:
$\mathbf{s} \odot \overline{\mathbf{b}} = \mathbf{a} \odot \cancel{\mathbf{b}  \odot \overline{\mathbf{b}}}  = \mathbf{a}$.

However, if none of the factors are given, then while decoding the vector is still feasible, it becomes a combinatorial search problem whose complexity grows exponentially with the number of factors. 
For instance, if there are 100 possibilities for factor $\mathbf{a}$ and 100 for $\mathbf{b}$, then the challenge is to search over all 10,000 combinations of two factors.
%This problem often occurs in symbolic manipulation problems, for example, finding the position of a given item in a tree structure~\cite{FradyResonator2020}.
Recent work~\cite{FradyResonator2020,KentResonatorNetworks2020} proposed an elegant mechanism, called the resonator network, to address the challenge of factorizing $\mathbf{s}$ into its arguments. 
%The resonator network, in essence, is a recurrent neural network that uses VSA principles to address the combinatorial search problem.

To factor the representation from the input hypervector $\mathbf{s}$, the resonator network uses multiple populations,  $\hat{\mathbf{a}}(t)$, and $\hat{\mathbf{b}}(t)$, each of which tries to infer a particular factor from the input hypervector. 
Each factor that is a possibility is stored in a separate factor item memory ($\mathbf{A}$, $\mathbf{B}$). 
Each population, called a resonator, communicates with the input hypervector and the other populations using the following dynamics:
\noindent
\begin{equation}
\begin{split}
&\mathbf{\hat{a}}(t+1)= f_n \Big( \mathbf{A} \Re \Big( \mathbf{A}^{\dagger} (\mathbf{s} \odot  \overline{\mathbf{\hat{b}}}(t)) \Big) \Big); \\
&\mathbf{\hat{b}}(t+1)= f_n \Big( \mathbf{B} \Re \Big( {\mathbf{B}}^{\dagger} (\mathbf{s} \odot  \overline{\mathbf{\hat{a}}}(t)) \Big) \Big). 
\end{split}
\label{eq:resnet:text}
\end{equation}
\noindent
% \noindent
% \begin{equation}
% \begin{split}
% &\mathbf{\hat{a}}(t+1)= f_n \Big( \mathbf{A} \Re \Big( \mathbf{A}^{\dagger} (\mathbf{s} \odot  \overline{\mathbf{\hat{b}}}(t)  \odot  \overline{\mathbf{\hat{c}}}(t) ) \Big) \Big); \\
% &\mathbf{\hat{b}}(t+1)= f_n \Big( \mathbf{B} \Re \Big( {\mathbf{B}}^{\dagger} (\mathbf{s} \odot  \overline{\mathbf{\hat{a}}}(t)  \odot  \overline{\mathbf{\hat{c}}}(t) ) \Big) \Big); \\
% &\mathbf{\hat{c}}(t+1)= f_n \Big( \mathbf{C} \Re \Big( {\mathbf{C}}^{\dagger} (\mathbf{s} \odot  \overline{\mathbf{\hat{a}}}(t)  \odot  \overline{\mathbf{\hat{b}}}(t) ) \Big) \Big).
% \end{split}
% \label{eq:resnet:text}
% \end{equation}
% \noindent
%% Previous version:
% \noindent
% \begin{equation}
% \begin{split}
% &\mathbf{\hat{a}}(t+1)= f_n \Big( \mathbf{A} \Re \Big( {\mathbf{A}}^{\top} \overline{(\mathbf{s} \odot  \overline{\mathbf{\hat{b}}}(t)  \odot  \overline{\mathbf{\hat{c}}}(t) )} \Big) \Big); \\
% &\mathbf{\hat{b}}(t+1)= f_n \Big( \mathbf{B} \Re \Big( {\mathbf{B}}^{\top} \overline{(\mathbf{s} \odot  \overline{\mathbf{\hat{a}}}(t)  \odot  \overline{\mathbf{\hat{c}}}(t) )} \Big) \Big); \\
% &\mathbf{\hat{c}}(t+1)= f_n \Big( \mathbf{C} \Re \Big( {\mathbf{C}}^{\top} \overline{(\mathbf{s} \odot  \overline{\mathbf{\hat{a}}}(t)  \odot  \overline{\mathbf{\hat{b}}}(t) )} \Big) \Big).
% \end{split}
% \label{eq:resnet:text}
% \end{equation}
% \noindent
%%

\begin{figure*}[t]%[H]%[!ht]%[t!]
\centering
\includegraphics[width=2.0\columnwidth]{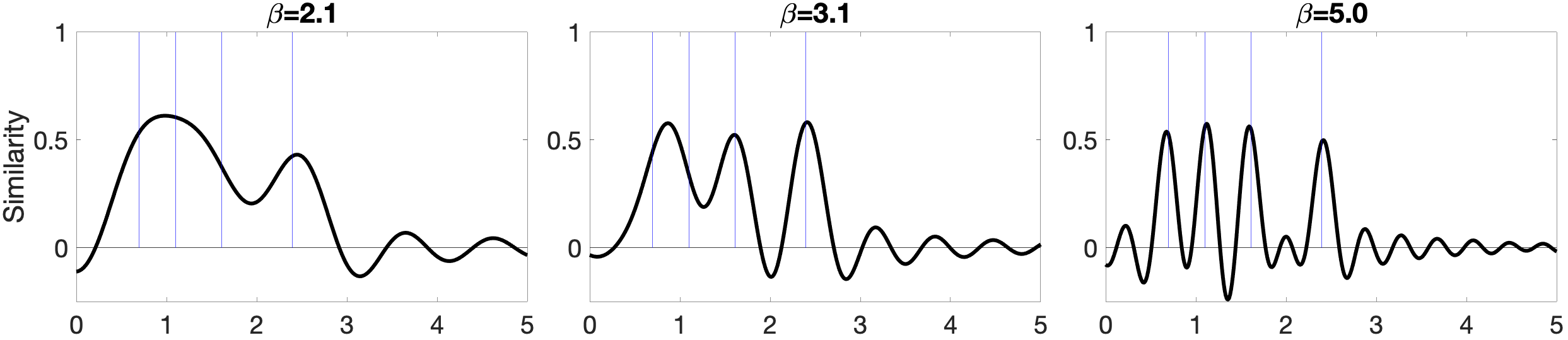}
\caption{
An example of varying behavior of the superposition of FPEs corresponding to a set of scalars: $\{\log(2), \log(3), \log(5), \log(11)\}$; $y$-axis depicts the average cosine similarity (thick lines) between the superposition hypervector and FPEs of scalars in $x$-axis.  
Thin vertical lines correspond to the locations of the elements of the considered set. 
The corresponding panels show cosine similarities for FPEs formed with different values of $\beta$.
The reported values are averages computed from $30$ simulation runs; $n=512$.
}
\label{fig:superposition}
\end{figure*}

The process is iterative and progresses in discrete time steps, $t$.
In essence at time $t$, each population can hold multiple weighted estimates for one of the factors through the VSA principle of superposition~\cite{FradyCapacity2018, KleykoPerceptron2020}. 
This allows a population to test multiple guesses for factor identity simultaneously. 
Each resonator uses the current estimates from the other populations to invert the input hypervector and infer the factor of interest.\footnote{ 
Note that in~(\ref{eq:resnet:text}) the update is synchronous, i.e., estimates at $t+1$ are based on the estimates from the previous $t$th iteration. 
It is possible to update estimates asynchronously.
We use this asynchronous mode to perform the evaluation (see Section~\ref{fact:setup}). 
}
The cost of superposition is crosstalk noise, making the inference step noisy when many estimates are tested at once. 
Therefore, the next step uses the factor item memory to remove the extraneous estimates. 
The estimate 
%$\mathbf{\hat{f}}_i$ % this is confusing 
for each factor is \emph{cleaned up} by constraining the resonator activity only to the allowed atomic hypervectors stored in the corresponding factor item memory.
Finally, a regularization step (denoted as $f_n(*)$) limiting the values of components of new estimates is needed.
Successive iterations of this inference and clean up procedure (\ref{eq:resnet:text}) eliminate the noise as the factors become identified and find their place in the input vector. 
When the factors are fully identified, the resonator network reaches a stable equilibrium, and the factors can be deduced from the stable activity pattern.

\section{Results}

\subsection{Logarithmic FPE for encoding integers}
\label{fact:setup}

We will use the problem of factorizing semiprimes (composite integers with exactly 2 prime factors) to demonstrate how FPE and resonator networks can interact together to solve this problem. 
A semiprime $s$ is obtained simply by multiplying two primes $x$ and $y$:
\noindent
\begin{equation}
\label{eq:semiprime} 
s=xy.
\end{equation}
\noindent

We use $\mathcal{P}(s)$ to denote the set of all primes that are potential factors of $s$ (since the minimum prime factor is two, all primes less than or equal to $s/2$ are possible factors). The $i$th element of the set is denoted as $\mathcal{P}(s)_i$.
When setting up the factorization search, the set $\mathcal{P}(s)$ should be known. 
The set is used to form the item memory $\mathbf{\Phi}$ containing hypervectors for all primes in $\mathcal{P}(s)$. 
The hypervector of $\mathcal{P}(s)_i$ (denoted as $\mathbf{\Phi}_i$) is formed with FPE (see Section~\ref{sec:fpe}) as:
%using the random base vector $\mathbf{z}$ as:
\noindent
\begin{equation}
    \mathbf{\Phi}_i = \mathbf{z}(\log(\mathcal{P}(s)_i)) = \mathbf{z}^{\beta \log(\mathcal{P}(s)_i)}.
    \label{eq:fpe:codebook}
\end{equation}
\noindent
Critically, the combination of the properties of the log transformation and FPE (see (\ref{eq:fpe:sum})) results in the following behavior of hypervectors:  \noindent
\begin{equation}
\begin{split}
    & \mathbf{z}(\log(s)) =  \mathbf{z}^{\beta \log(s)} = \mathbf{z}^{\beta \log(xy)}  
     =  \mathbf{z}^{\beta (\log(x) + \log(y) )}\\
     &  =  \mathbf{z}^{\beta \log(x)}   \odot \mathbf{z}^{\beta \log(y)}  
     = \mathbf{z}(\log(x)) \odot \mathbf{z}(\log(y)).
\end{split}
\end{equation}
\noindent
In other words, the log transformation combined with FPE results in the binding between vectors being equivalent to the FPE vector of the product. 
This allows one to express the problem of semiprime factorization in terms of vector factorization, which can be solved through the resonator network formulation as in Section~\ref{sect:fac:rn}, where each resonator uses the codebook $\mathbf{\Phi}$.

\subsection{FPEs in superposition}
\label{sec:fpe:pecul}

Traditionally in VSA, the most straightforward use of the superposition operation is to represent a set of elements~\cite{KanervaHyperdimensional2009,KleykoABF2020}.
As with any other VSA representation, FPEs can also be used with the superposition operation to, e.g., represent a set of scalar values.  
However, due to the similarity-preserving properties of FPEs, the hypervector resulting from the superposition of several FPEs might exhibit counter-intuitive ``hybrid'' behavior, which is neither fully symbolic nor fully subsymbolic.
When utilizing the resonator network to solve the factorization problem, we need to account for the behavior of FPE superpositions.

Consider an example of the following set: \\ $\{\log(2)$, $\log(3)$, $\log(5)$, $\log(11)\}$, where the hypervector $\mathbf{s}$ representing the set is formed using the FPEs corresponding to the elements of the set as: 
\noindent
\begin{equation}
\label{eq:set} 
\mathbf{s} = \mathbf{z}^{\beta \log(2)} + \mathbf{z}^{\beta \log(3)} +\mathbf{z}^{\beta \log(5)} +\mathbf{z}^{\beta \log(11)}.
\end{equation}
Fig.~\ref{fig:superposition} presents the average cosine similarity between $\mathbf{s}$ and the corresponding FPEs of scalars along the $x$-axis for different values of $\beta$.
%(panels, $\beta \in \{2.1, 3.1, 5.0 \}$). 
The first thing to notice in Fig.~\ref{fig:superposition} is that the choice of $\beta$ profoundly affects the obtained similarity distributions.  
%from Fig.~\ref{fig:bandwidth} 
%Lower values of $\beta$ imply wider similarity kernels. 
%This explains why the left panel in Fig.~\ref{fig:superposition} with the lowest value of $\beta$ provides only two peaks of similarity: one at $\log(11)$ and one around the remaining values of the set. 
When $\beta=2.1$, FPEs of $\{\log(2), \log(3), \log(5)\}$ are similar to each other, and so when superimposed together, they interact in such a way that there is one large peak of similarity near their mean.
Increasing $\beta$ to $3.1$ splits the large peak into two: one for $\log(5)$ and one in between $\log(2)$ and $\log(3)$. 
This is intuitive since $\log(2)$ and $\log(3)$ are closer to each other than $\log(3)$ and $\log(5)$.
Finally, once $\beta$ is large enough (e.g., $\beta=5.0$) all four peaks become clearly distinct and they correspond to the values of the elements in the set.
%, and the elements of the set can be cleanly identified. 

\begin{figure*}[t]%[H]%[!ht]%[t!]
%\begin{wrapfigure}[9]{R}{0.5\linewidth} 
%\vspace{-3ex}
\begin{minipage}[h]{0.35\linewidth}
\center{\includegraphics[width=1.0\linewidth]{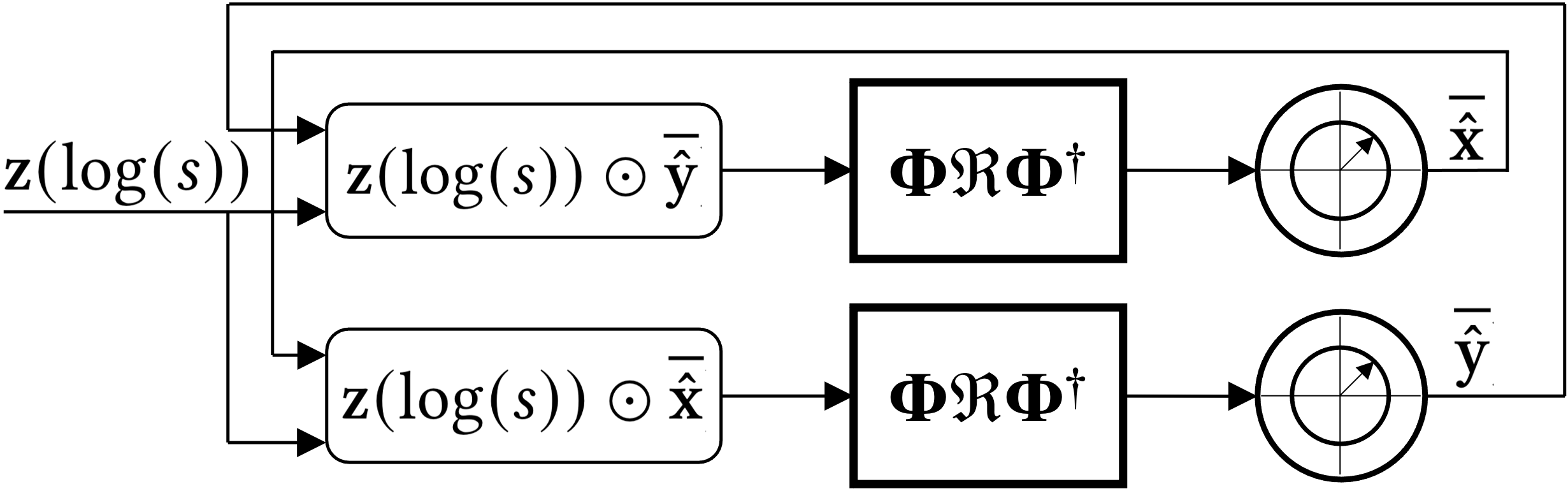} \\ A}
\end{minipage}
\hfill
\begin{minipage}[h]{0.64\linewidth}
\center{\includegraphics[width=1.0\linewidth]{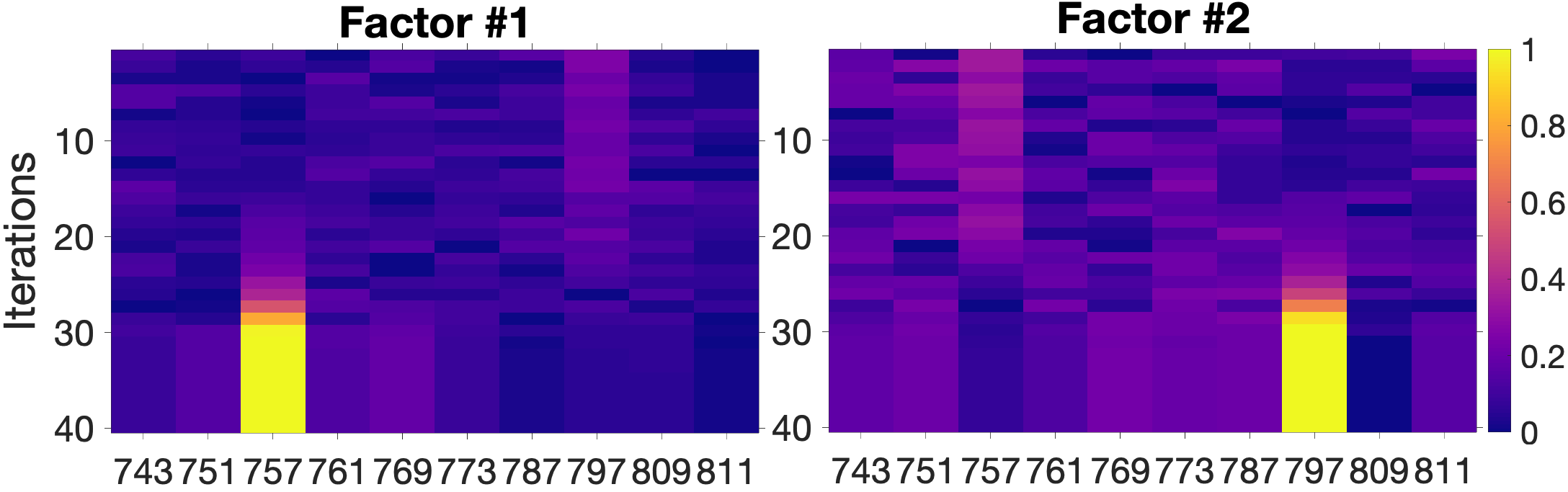} \\ B}
\end{minipage}
%\centering
%\includegraphics[width=\linewidth]{img/resonator_network}
\caption{
Left panel (A): an example of a resonator network with two factors for integer factorization according to~(\ref{eq:resnet:semiprimes}).
Each resonator uses the estimate from the other resonator to infer one of the factors (e.g. $\mathbf{z}(\log(s)) \odot \overline{\hat{\mathbf{y}}}$), these estimates are then cleaned-up by limiting them to the span of the codebook ($\mathbf{\Phi} \Re \mathbf{\Phi}^\dagger$), and finally the vector elements are restored to unit magnitude phasors ($f_n(x_i) = x_i/|x_i|$).
%The diagram presents normalization of components to the unit magnitude in place of the nonlinear function.
Right panel (B): an example of convergence of a resonator network. Color values correspond to the normalized inner product between $\mathbf{\hat{x}}$ (Factor \# 1) and $\mathbf{\hat{y}}$ (Factor \# 2) and the entries of the codebook $\mathbf{\Phi}$ corresponding to the primes depicted on $x$-axis. A small range of primes are shown for visualization. The dynamics are initially very chaotic until around iteration 30 where the network identifies the solution and quickly reaches a stable equilibrium.
}
\label{fig:resonator}
%\end{wrapfigure}
\end{figure*}

Thus, setting the value of $\beta$ allows traversing between two extremes: when $\beta$ is very small, FPEs of scalars that are far away from each other are still very similar (subsymbolic behavior) while when $\beta$ is very large, FPEs of scalars that are near each other are dissimilar (symbolic behavior).
%One way to look at this behavior is from point of view where the sinc function acts as a band-limited delta function. Since resolving nearby scalars requires high-frequency changes (i.e. a ``dip'' between two nearby peaks in Fig.~\ref{fig:superposition}) the band-limit of the sinc function explains the merging.
In other contexts, we expect that different applications might favor different modes.
For example, when working with clustering problems~\cite{BandaragodaTrajectoryTraffic2019, ImaniHDCluster2019,KleykoBoostingSOM2019,  HernandezClustering2021, OsipovHyperSeed2021}, there is potential that subsymbolic merging would be useful for generating centroids and computing means.
On the other hand, for integer factorization, it is important to choose $\beta$ such that we only operate in the symbolic mode, as we desire that each different value is treated as a distinct alternative within the solution space of the factorization problem. 
Practically, this means we have chosen $\beta$ to be large so that the inner product between the hypervectors of any two adjacent primes would be close to zero.
This was achieved by setting $\beta$ as :
\begin{equation}
\beta=\frac{10^4}{\min_i(\log(\mathcal{P}(s)_{i+1}) - \log(\mathcal{P}(s)_{i})) }
\label{eqn:beta}
\end{equation}
The extra factor of $10^{4}$ was added to ensure that $\beta$ was always sufficiently large. 
% We include experiments examining the affect of $\beta$.

\subsection{Factorization of semiprimes with the resonator network}

In this case, the dynamics of the resonator network factorizing $s$ into $x$ and $y$ is described as follows: 
%% Changing here
% I'm pretty sure that these are actually the same mathematically, especially since the real part step is included. I think this is actually the conjugation compared to previous version, but because of real part step they end up the same.
\noindent
\begin{equation}
\begin{split}
&\mathbf{\hat{x}}(t+1)= f_n \Big( \mathbf{\Phi} \Re \Big( {\mathbf{\Phi}}^{\dagger} (\mathbf{z}(\log(s)) \odot  \overline{\mathbf{\hat{y}}}(t)) \Big) \Big); \\
&\mathbf{\hat{y}}(t+1)= f_n \Big( \mathbf{\Phi} \Re \Big( {\mathbf{\Phi}}^{\dagger} (\mathbf{z}(\log(s)) \odot  \overline{\mathbf{\hat{x}}}(t+1)  ) \Big) \Big), 
\end{split}
\label{eq:resnet:semiprimes}
\end{equation}
\noindent
% Previous version:
% \noindent
% \begin{equation}
% \begin{split}
% &\mathbf{\hat{x}}(t+1)= f_n \Big( \mathbf{\Phi} \Re \Big( {\mathbf{\Phi}}^{\top} \overline{(\mathbf{z}(\log(s)) \odot  \overline{\mathbf{\hat{y}}}(t))} \Big) \Big); \\
% &\mathbf{\hat{y}}(t+1)= f_n \Big( \mathbf{\Phi} \Re \Big( {\mathbf{\Phi}}^{\top} \overline{(\mathbf{z}(\log(s)) \odot  \overline{\mathbf{\hat{x}}}(t+1)  )} \Big) \Big), 
% \end{split}
% \label{eq:resnet:semiprimes}
% \end{equation}
% \noindent
where $\hat{\mathbf{x}}(t)$ and $\hat{\mathbf{y}}(t)$ denote the hypervectors corresponding to the current estimates of the resonator network for $\mathbf{z}(\log(x))$ and  $\mathbf{z}(\log(y))$;
$f_n(x_i) = x_i/|x_i|$ normalizes each component to unit magnitude.
Note that in (\ref{eq:resnet:semiprimes}) both factors use the same item memory $\mathbf{\Phi}$ since both $x$ and $y$ are present in $\mathcal{P}(s)$.
Once the resonator network converges or reaches the maximum number of iterations, the most recent estimates of the resonator network are used to obtain the predictions $\hat{x}$ and $\hat{y}$ corresponding to the primes in $\mathcal{P}(s)$ whose hypervectors in $\mathbf{\Phi}$ are the most similar to $\hat{\mathbf{x}}(t)$ and $\hat{\mathbf{y}}(t)$. 
In the experiments, the maximum number of iterations was $100$.
A schematic overview of the resonator is shown in Fig.~\ref{fig:resonator}.

Let us walk-through an example of the factorization process described above. 
Assume that we would like to factorize semiprime $s=603,329$ into its factors ($x=757$ and $y=797$). 
First, we need to define all the primes in $\mathcal{P}(s)$ that will be used to form the item memory $\mathbf{\Phi}$.
%For the sake of simplicity, we can assume $\mathcal{P}(s)$ contains all primes that are less than or equal to $s/2$.
In this case, the cardinality of $\mathcal{P}(s)$ will be $|\mathcal{P}(s)|=26,135$ with the smallest prime being $2$ and the largest one being $301,657$.
Once $\mathcal{P}(s)$ is fixed, we can use our equation for $\beta$ (\ref{eqn:beta}) to calculate $\beta\approx\num{1.5e9}$.
Next, we need to choose a suitable dimensionality of hypervectors $n$ (see the next section for performance evalutation of different values of $n$).
Then, a random $n$-dimensional base vector $\mathbf{z}$ is generated. 
The base vector $\mathbf{z}$ is used to populate the item memory $\mathbf{\Phi}$ with hypervectors corresponding to FPEs of logarithms of elements of $\mathcal{P}(s)$ according to (\ref{eq:fpe:codebook}).  
We also form the FPE of the given semiprime $s$ as $\mathbf{z}(\log(s)) = \mathbf{z}^{\beta \log(s)}$.
The final step is to setup and run the resonator network according to~(\ref{eq:resnet:semiprimes}) using the obtained $\mathbf{z}(\log(s))$ and $\mathbf{\Phi}$.
The initial estimates for  $\hat{\mathbf{x}}(0)$ and $\hat{\mathbf{y}}(0)$ are set to the normalized superposition of all hypervectors in $\mathbf{\Phi}$.
If $n$ is large enough, then after several iterations with high probability the resonator network will converge. The final state of $\hat{\mathbf{x}}(t)$ and $\hat{\mathbf{y}}(t)$ can be matched to the closest hypervectors in $\mathbf{\Phi}$, which will correspond to primes $757$ and $797$ (Fig. \ref{fig:resonator}). Note that in this configuration, $\mathbf{\hat{x}}$ and $\mathbf{\hat{y}}$ can converge to either one of the primes.

\begin{figure*}[tb]%[!ht]%[t!]
\centering
\includegraphics[width=1.99\columnwidth]{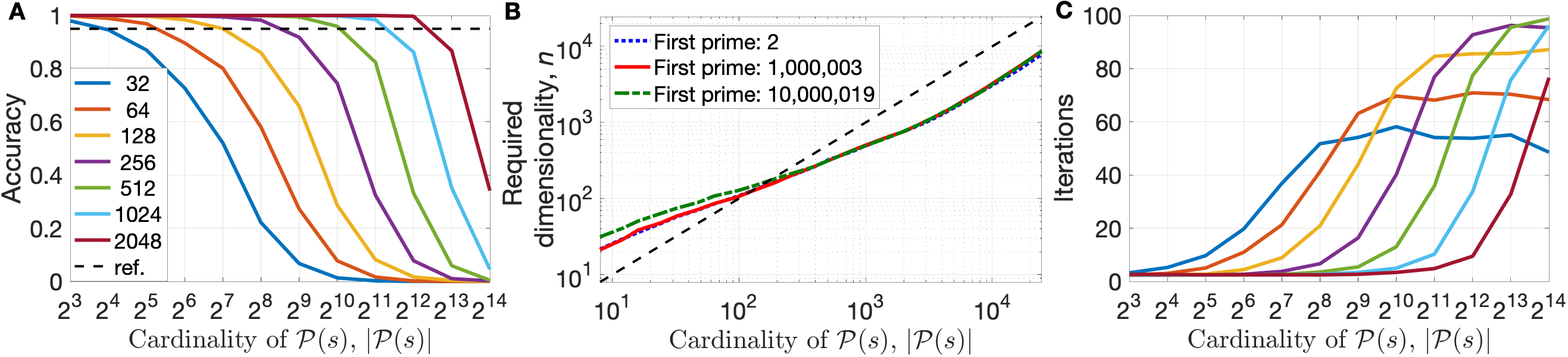}
\caption{
Left panel (A): the accuracy of semiprimes factorization against the cardinality of $\mathcal{P}(s)$.
The reported values are averages computed from $4,000$ random semiprimes.
Central panel (B): 
average minimal dimensionality of hypervectors ($y$-axis) required to achieve at least 95\% successful factorization for the given cardinality of $\mathcal{P}(s)$ ($x$-axis). 
Thin black dashed line depicts the linear relation between $n$ and $|\mathcal{P}(s)|$.
The colored lines correspond to different starting points used to form $\mathcal{P}(s)$.
%Note that solid and dash-dotted lines coincide.
The reported values are averages computed from $10$ simulation runs.
During each simulation run, $1,000$ randomly chosen semiprimes were used to assess the factorization accuracy for every considered value of $n$.
Right panel (C): number of iterations used by the resonator network to either converge to a solution or to reach the maximum number of iterations (set to $100$).
%The stopping condition was that the cosine similarity between predictions at neighboring time steps for all factors was greater than or equal to $0.999$.
%The reported values are averages computed from $4,000$ random semiprimes.
}
\label{fig:fact:fixed:dim}
\end{figure*}

%\newpage
\subsection{Empirical evaluation of performance and scaling }
\label{sect:empirical}

In this section, we report the results of the empirical evaluation of the proposed approach. 
In the experiments below, we need to measure the success of the factorization by the average accuracy of correctly factorizing many semiprimes $s=xy$ (where $x$ and $y$ are chosen randomly from $\mathcal{P}(s)$).

First, we examine the factorization accuracy against the number of elements in $\mathcal{P}(s)$ and as a function of dimensionality, reported in left panel in Fig.~\ref{fig:fact:fixed:dim}.
%for $n \in \{32, 64, 128, 256, 512, 1024, 2048\}$. 
For each value of $n$, we can identify three regimes with respect to the accuracy:  
high-fidelity, where the accuracy is nearly perfect, low-fidelity, where the accuracy is not perfect but above chance, and random guessing, where the accuracy is effectively chance. 
It is evident that with increased $n$, the maximum size of $\mathcal{P}(s)$ within the high-fidelity regime also increased. 
% this is on a log scale so you cant tell, its not linear its quadratic! 
%Roughly speaking, the increase seems linear but, in fact, it scaled slightly better. For example, for $n=32$ and $|\mathcal{P}(s)|=32$, the accuracy was $0.870$, while for $n=2,048$ and $|\mathcal{P}(s)|=4,096$, the accuracy was still $0.997$. 

It is also important to estimate how the complexity of the approach scales. To do so, we consider the dimensionality of hypervectors required to perform the factorization successfully for the given cardinality of $\mathcal{P}(s)$.
%For the experiment below, 
We have defined the successful factorization as the accuracy that is greater than or equal to $0.95$.
The scaling of the required dimensionality of hypervectors converges to a line with slope of approximately 1 with respect to cardinality of $\mathcal{P}(s)$ (central panel in Fig. \ref{fig:fact:fixed:dim}). 
This observation is in line with the experiments reported in~\cite{KentResonatorNetworks2020}, where random hypervectors were used to form an abstract factorization problem. 
Practically, this also means that for large $|\mathcal{P}(s)|$, reasonable values of $n$ would be sufficient to perform the factorization. 

Recall that in Section~\ref{fact:setup}, we discussed the choice of suitable $\beta$ for a given $\mathcal{P}(s)$ to continue operating in the symbolic mode.
Intuitively, the potential issue with scaling $\beta$ is that when  $\mathcal{P}(s)$ contains large primes, $\beta$ will also be large due to the use of the log transformation. Large values of $\beta$ could cause numerical issues when performing the FPE. In order to demonstrate the potential role of $\beta$ on the factorization performance, Fig.~\ref{fig:fact:fixed:dim}B also depicts the required dimensionalities of hypervectors for primes in $\mathcal{P}(s)$ that were picked using different starting points, which following (\ref{eqn:beta}) leads to different values for $\beta$. 
First, the results suggest that the ranges of primes requiring the use of larger values of $\beta$ did not incur a drastic increase in dimensionality of hypervectors, so factorization performance is mainly limited by the capacity of the resonator.
Second, since larger values of $\beta$ are not an issue, the proposed approach can handle varying ranges of primes.

\begin{figure*}[tb]%[!ht]%[t!]
\centering
\includegraphics[width=1.99\columnwidth]{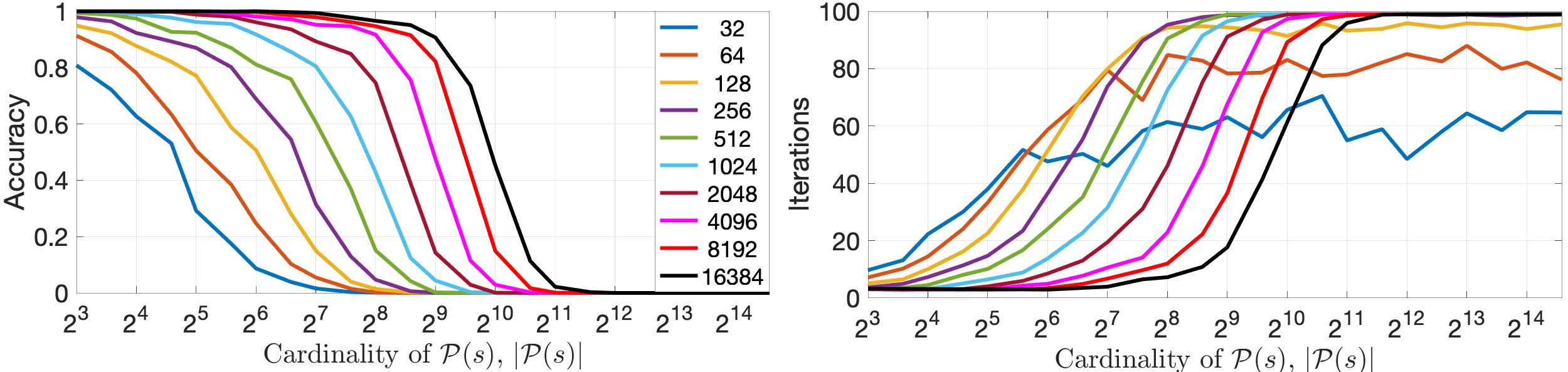}
\caption{
Left panel: the accuracy of $3$-almost primes factorization against the cardinality of $\mathcal{P}(s)$.
Right panel: number of iterations used by the resonator network to either converge to a solution or to reach the maximum number of iterations (set to $100$).
%The stopping condition was that the cosine similarity between predictions at neighboring time steps  for all factors was greater than or equal to $0.999$.
%
The values are averages from $1,000$ randomly chosen $3$-almost primes.
}
\label{fig:fact:fixed:dim:3Factors}
\end{figure*}

In addition to the factorization accuracy, it is also worth looking at the average number of iterations used by the resonator network to converge (right panel in Fig.~\ref{fig:fact:fixed:dim}). 
There is a clear correspondence between the accuracy and the number of iterations of the resonator network. 
When the resonator network was in the high-fidelity regime, only a few iterations were required to find a solution, and $|\mathcal{P}(s)|$ increased the number of iterations also increased. 
Finally, once $|\mathcal{P}(s)|$ was too large for a chosen $n$, the resonator converged to the wrong answer or reached the iteration limit.

\begin{comment}
%\newpage
\begin{figure}[tb]%[!ht]%[t!]
\centering
\includegraphics[width=1.0\columnwidth]{img/Required_Dim_2_factors}
\caption{
Average minimal dimensionality of hypervectors ($y$-axis) required to achieve at least 95\% successful factorization for the given cardinality of $\mathcal{P}(s)$ ($x$-axis); $|\mathcal{P}(s)| \in~2^{[3:0.5:14.5]}$. 
Thin black dashed line depicts the linear relation between $n$ and $|\mathcal{P}(s)|$.
The colored lines correspond to different starting points used to form $\mathcal{P}(s)$.
%Note that solid and dash-dotted lines coincide.
The reported values are averages computed from $10$ simulation runs.
During each simulation run, $1,000$ randomly chosen semiprimes were used to assess the factorization accuracy for every considered value of $n$.
%Thin dashed line is an indicator of a perfect linear (with slope $1$) dependency between $n$ and $|\mathcal{P}(s)|$. 
}
\label{fig:fact:scaling}
\end{figure}
\end{comment}

\subsection{Factoring composite numbers beyond semiprimes}
%Fig.~\ref{fig:fact:scaling} reports the results ...
The proposed approach is not limited to semiprimes. 
In principle, it can be applied on any $k$-almost prime.
In Fig.~\ref{fig:fact:fixed:dim:3Factors}, we report the case of factorization of integers with three factors ($s=abc$) using a similar setup as above. Now there are three resonators, and each resonator is designed for three factors. They each have a similar update dynamics, for instance for the first factor:
\begin{equation}
\mathbf{\hat{a}}(t+1)= f_n \Big( \mathbf{\Phi} \Re \Big( {\mathbf{\Phi}}^{\dagger} (\mathbf{z}(\log(s)) \odot  \overline{\mathbf{\hat{b}}}(t)) \odot \overline{\mathbf{\hat{c}}}(t) \Big) \Big).
\end{equation}
The observed results are consistent with that obtained for the case of semiprimes in Fig.~\ref{fig:fact:fixed:dim}. 
Note that compared to the case of semiprimes, for $3$-almost primes larger values of $n$ were required to get to the high-fidelity regime. This is expected since for the semiprimes the search space grows as $|\mathcal{P}(s)|^2$ while for $3$-almost primes it grows much faster as $|\mathcal{P}(s)|^3$.
The approach can be extended to other composite numbers by including more resonators in the network. Again, the capacity and likelihood of solving the factorization problem depend on the combinatorics of the factors, and this grows exponentially with number of factors. Further, the identity vector (vector of all $1$s, a.k.a. $\mathbf{z}(\log(1)) = \mathbf{z}^{0}$) can be added to $\mathbf{\Phi}$ to enable solving problems with unknown number of factors.

\section{Discussion}
\label{sect:discussion}

\subsection{Summary of the study}

Our goal was to demonstrate that while Vector Symbolic Architectures (VSA)~\cite{KanervaHyperdimensional2009, KleykoComputingParadigm2021} were originally proposed to solve problems in cognition, VSAs are a highly flexible and powerful framework for expressing challenging computational problems in high-dimensional vector spaces.
%We focused on demonstrating the reasoning on real-valued data using compositional distributed representations within the framework of Vector Symbolic Architectures (VSA)~\cite{KanervaHyperdimensional2009, KleykoComputingParadigm2021}.
We used integer factorization to showcase both the expressiveness of VSAs and novel techniques of representing numbers in high-dimensional vectors.
VSAs are now well-known as frameworks for many novel computational devices that are designed for highly efficient and parallel computations \cite{rahimi2007random,RahimiNanoscalable2017,DaviesAdvancingLoihi2021}.
Using VSAs to express challenging computational problems that can be solved by neural network architectures, like the resonator network, brings out the potential of utilizing neuromorphic hardware. It is relatively straight-forward to scale VSA algorithms like the resonator network -- this simply means expanding the dimensionality of the vector representations. This ease of scalability is compatible with large scale meshes of neuromorphic chips \cite{NNkNN20}. While we did not execute these experiments on neuromorphic hardware, there are several previous models \cite{EliasmithSPAUN2012} and novel proposals \cite{FradyTPAM2019, FradySDR2020} that neuromorphically perform VSA computations.

One of our main contributions was expanding our understanding of how number systems can be expressed in vector spaces. 
Recently the technique of fractional power encoding (FPE) has been gaining new attention as a way to represent geometrical spaces, maps, manifolds and functions~\cite{PlateNested1994,KomerContinuous2019, FradyFunctions2021, FradyFunctionsNICE2022}. 
Integers are the ordinal data type, and, therefore, their distributed representation should preserve the data topology, which provides a proper setup for the use of fractional power encoding.
Previously integers were easily represented by the integer powers of an FPE, but in this formulation the binding operation results in the FPE of the integer sum.
%Integer factorization involves scalar multiplication that is equivalent to addition after log transformation, which in turn corresponds to the binding of the respective fractional power encodings. 
By expressing the FPE with the logarithms of integers, we enabled a representation of integer values where the binding operation now leads to the FPE of the integer product.
With this formulation for representing integers, we could then express the integer factorization problem as the problem of vector factorization, which can be solved using resonator networks \cite{FradyResonator2020}.

This FPE representation of logarithmic integers meant we needed to examine the consequences of superpositions of FPE vectors. The resonator network uses the principle of \emph{search in superposition}, and for it to successfully solve the factorization problem, the individual factors need to be uniquely identifiable by their FPE vectors. 
We used the parameter $\beta$ to rescale the FPE vectors so that the logarithmic FPE representations were sufficiently spaced such that their similarity kernels were not overlapping.
There is a simple strategy for scaling $\beta$ based on the minimum log distance between neighboring primes, but we also showed that there are few side-effects for making $\beta$ much larger, and generally factorization performance is not too dependent on $\beta$ as long  as it is sufficiently large. 

% From our point of view, there are several advantages with focusing on the application of VSAs to integer factorization. %in place of the reasoning problem.
% First, integer factorization is a well-known and intuitive problem that does not require lengthy introduction.  
% Second,
% Third, the 
% Finally, 
% the fact that the problem can be formulated as the binding of distributed representation of the factors allows using the resonator network~\cite{FradyResonator2020,KentResonatorNetworks2020} to find out the factors (i.e., primes) for a given compositional entity (i.e., a semiprime).
% Thus, integer factorization allows demonstration of several VSA primitives: binding operation, fractional power encoding, resonator network, and superposition operation (as a part of a resonator network).

The way the factorization problem is solved by the resonator network is best described via the concept of \emph{search in superposition}. During this process, many number combinations may be considered simultaneously, something that is not possible with conventional digital number representations.  
We believe that the extended idea of \emph{computing in superposition} ~\cite{KleykoComputingParadigm2021} is a particularly important aspect of VSAs that should be investigated further.
It is also worth emphasizing, that the resonator network can be used beyond semiprimes (i.e., with more than two factors; see Section~\ref{sect:empirical}). 

% The empirical evaluation investigated two potential factors that could affect performance, the scale of the resonator network and the FPE parameter $\beta$. 
% The experiments with $\beta$
%  the former plays a more important role since comparable dimensionalities of hypervectors were required to successfully factorize semiprimes formed from much larger primes. 

%\todo[inline, color=green]{
%What else to be said here? 
%}

\subsection{Related work}
\label{sect:disc:realted}

%It is important to note that rather than claiming the superiority of the proposed approach for semiprimes factorization, the goal of this paper is to demonstrate how a well-known problem of semiprimes factorization  can be formulated in terms of VSA primitives such that the resonator network can be used to solve the problem.  

Our main contribution is a formalism for using VSAs to solve integer factorization, providing a way to solve classical factorization problems with distributed representations. 
Though our formalism benefits from the properties of VSAs (such as being distributed, robust, and computing in superposition), it was not our goal to demonstrate the superiority to other algorithms.
The promise of our approach is really the potential of scalable and efficient execution of such an algorithm on neuromorphic hardware.
It is important to keep in mind that the proposed approach should not be considered as a panacea in terms of providing a straight-forward polynomial solution to the semiprimes factorization problem. 
The number of primes grows exponentially w.r.t. the number of bits used to represent a number, which also implies the exponential growth of the resonator.

%Nevertheless, it is important to contrast our approach to alternative non-VSA approaches to optimization and factorization problems. 
%Nevertheless, we contrast our proposed approach to other similar but non-VSA approaches."
%Nevertheless, it is important to contrast the proposed approach to alternative ways of approaching the optimization and factorization problems. 

This work expands on previous efforts to solve optimization and factorization problems with neural attractor networks and physics-based architectures. 
Early work proposed to design neural attractor networks based on matrix-type auto-associative memories~\cite{palm1980associative, hopfield1982neural, FrolovWillshaw2002, FrolovTime2006, knoblauch2010memory} so that their dynamics is governed by a Lyapunov function that represents the objective of a particular optimization problem, for example, the path length in a traveling-salesman problem~\cite{hopfield1985neural}. The fixed point attractor dynamics of such networks searches the solution space and settles at an approximate solution.
%investigated the use of distributed representations in combinatorial optimization problems by searching for  to instances of  by some modifications of matrix-type auto-associative memories~\cite{palm1980associative, hopfield1982neural, FrolovWillshaw2002, FrolovTime2006, knoblauch2010memory,GritsenkoAMSurvey2017}. 
The resonator network similarly settles at a solution if there is one. However, its dynamics is not governed by an energy landscape/Hamiltonian, if there is no solution, it can converge to limit cycles or chaotic orbits. It has been shown empirically that this richer dynamic repertoire accelerates the search and, as a result, outperforms gradient-based optimization significantly \cite{KentResonatorNetworks2020}.
%Our work builds on previous efforts by using resonator networks for solving integer factorization, a hard combinatorial optimization problem.
%
%Additionally, physics-based solvers typically map discrete variables in an optimization problem to physical bits which take discrete values and can be quantum (qubits) or stochastic ($p$-bits). 
%In contrast with previous methods, our work is unique in that variable representations in the optimization problem are distributed across many units in the VSA system.

Quantum computing has been proposed as a physics-based method for solving combinatorial optimization problems~\cite{apolloni1989quantum, shor1999polynomial}.
There are some similarities between quantum algorithms and the working principles of the resonator network used in this paper. 
The resonator network operates on sums of complex numbers to solve factorization problems, which can be regarded a classical analog of the superposition principle used in quantum computing.
Adiabatic quantum annealing~\cite{apolloni1989quantum} methods solve an optimization problem by using the quantum tunneling effect. The optimization problem can be mapped to the Hamiltonian of a quantum system. The optimal solution or global minimum of the problem objective is found by slowly evolving the potential from an initial easy Hamiltonian to a more complicated Hamiltonian.
%Jiang et al. \cite{jiang2018quantum} used the D-Wave 2000Q quantum computer to factor integers up to $376,289$. 
%Future work should examine the relationship between tunneling and resonator dynamics. 
%
Due to the challenge of building reliable and large quantum computers, there is a renewed interest in classical physics-based solvers, such as networks of coupled oscillators~\cite{wang2019oim, ahmed2021probabilistic}.
The variables in the described resonator networks are complex-valued phasors, which can be represented by oscillators or by spiking neural networks~\cite{FradyTPAM2019}.  

\subsection{Future work}
\label{sect:disc:future}

% The techniques used in this paper can be extended to potentially solve many new types of computational challenges using distributed representations. 
% Integer factorization is among many in a large class of well-known challenging computational problems that require extensive computational resources. 

% The flexibility of VSA to express these problems allows for new distributed neural network designs that can solve them potentially on novel computational devices. 

There are several extensions of our approach, including subsequent analyses, that seem especially promising: 
% Yeah but we did this here now so maybe cut this
%we can consider the more general case, where the number of prime factors is at most $k$. For example, simulations with exactly 3 prime factors were already briefly considered in Section~\ref{sect:empirical}.
% Or more general how to map and optimize for neuromorphic hardware. What are current challenges in implementations? cite specific hardware, what hardware is noisy? 
While we presented the algorithmic approach and its realization on the conventional parallel hardware (GPU), the real promise is in the implementing VSAs in neuromorphic hardware. 
To this date, it is still an open question how to implement a VSA system in such hardware in full.  
Probably the closest mapping to spiking hardware is provided via the Neural Engineering Framework~\cite{EliasmithNeural2003, BekolayNengo2014}, although the potential challenge of this approach is spike efficiency. 
An alternative mapping proposal that is spike efficient is via representing FHRR phasors with spike times~\cite{FradyTPAM2019}, but this approach is not yet extended to account for all VSA operations.  
Another promising hardware direction is in-memory computing~\cite{KarunaratneInMemory2020, KarunaratneHDAugmented2021}. 
Since the above hardware is inherently noisy, it would provide a natural setup for demonstrating the robustness of our proposed factorization approach to noise (as expected from simulations performed in \cite{KentResonatorNetworks2020}).
%
% This is a good point but expand it more. What are challenges for other problems? how might our techniques work in other challenges?
Another important direction is to design mapping of other difficult problems, such as the subset-sum problem and other combinatorial optimization problems, to resonator networks with FHRR. 
A particular challenge for designing mappings for other problems is the absence of strict guidelines directing mapping development, so for each problem the mapping has to be done ad hoc. %  
Another limitation is that although resonator networks are well-suited for finding exact solutions (i.e., solving equality problems), it is less obvious how to formulate the problem of finding a maximum or minimum.

%\section{Conclusion}
%\label{sect:conclusions}

%\todo[inline, color=green]{
%Do we need any formal conclusion? 
%}


\begin{acks}
FTS, BAO, CB, and DK  were supported by Intel's THWAI.
BAO and DK  were supported by AFOSR FA9550-19-1-0241.
DK was supported by the MSCA Fellowship (grant 839179). 
CJK was supported by the DoD through the NDSEG Fellowship.
FTS was supported by Intel and NIH R01-EB026955.
\end{acks}

%\newpage

\bibliographystyle{ACM-Reference-Format}
\bibliography{Bibliography}

%\newpage
%\appendix
%\section{Factorization of 3-almost primes}
%\label{app:3:factors}

\balance

\end{document}